\documentclass[a4paper,conference]{IEEEtran}
\usepackage{cite}
\usepackage{times}
\usepackage{graphicx}
\usepackage{amsmath,amssymb,amsfonts}
\usepackage{textcomp}
\usepackage{xcolor}

\usepackage{algorithmic}
\usepackage{algorithm}
\usepackage{subcaption}
\usepackage{color, soul}
\usepackage{booktabs}
\usepackage{balance}

\newlength\myindent
\newcommand\bindent{
	\begingroup
	\setlength{\itemindent}{\myindent}
	\addtolength{\algorithmicindent}{\myindent}
}
\newcommand\eindent{\endgroup}

\newcommand{\etal}{{\textit{et al}}.\@ }

\begin{document}
%
\title{Feature-Dependent Cross-Connections in Multi-Path Neural Networks}

%
\author{\IEEEauthorblockN{Dumindu Tissera\IEEEauthorrefmark{1}\IEEEauthorrefmark{2},
Kasun Vithanage\IEEEauthorrefmark{2},
Rukshan Wijesinghe\IEEEauthorrefmark{1}\IEEEauthorrefmark{2},
Kumara Kahatapitiya\IEEEauthorrefmark{1},\\
Subha Fernando\IEEEauthorrefmark{2} and
Ranga Rodrigo\IEEEauthorrefmark{1}}

\IEEEauthorblockA{\IEEEauthorrefmark{1}Department of Electronic and Telecommunication Engineering,
University of Moratuwa, Sri Lanka}
\IEEEauthorblockA{\IEEEauthorrefmark{2}CODEGEN QBITS Lab, University of Moratuwa, Sri Lanka}}

\maketitle

\begin{abstract}
Learning a particular task from a dataset, samples in which originate from diverse contexts, is challenging, and usually addressed by deepening or widening standard neural networks. As opposed to conventional network widening, multi-path architectures restrict the quadratic increment of complexity to a linear scale. However, existing multi-column/path networks or model ensembling methods do not consider any feature-dependent allocation of parallel resources, and therefore, tend to learn redundant features. Given a layer in a multi-path network, if we restrict each path to learn a context-specific set of features and introduce a mechanism to intelligently allocate incoming feature maps to such paths, each path can specialize in a certain context, reducing the redundancy and improving the quality of extracted features. This eventually leads to better-optimized usage of parallel resources. To do this, we propose inserting feature-dependent cross-connections between parallel sets of feature maps in successive layers. The weighting coefficients of these cross-connections are computed from the input features of the particular layer. Our multi-path networks show improved image recognition accuracy at a similar complexity compared to conventional and state-of-the-art methods for deepening, widening and adaptive feature extracting, in both small and large scale datasets.
\end{abstract}

%
\IEEEpeerreviewmaketitle

\section{Introduction}
\label{se:intro}
Learning a particular task in a dataset while handling the diversity among input samples, essentially requires the model to adapt to the context of the input. The naive approaches of network deepening or widening \cite{alexnet, vgg, resnet, wideresnet} tend to improve the performance as a result of the deep abstract feature extraction or the increased number of features extracted. Having multiple parallel paths/columns in each layer \cite{inception, resnetxt}, as opposed to conventional widening, prevents the quadratic increment of network complexity as a result of width enhancement. In model ensembling methods \cite{alexnet,  vgg}, each model is expected to converge to slightly different local optima, giving a better-combined performance. However, these approaches lack the adaptability to diverse contexts of input samples, and hence, any resource increment due to network deepening or widening is not well utilized and subjected to feature redundancy.  

\begin{figure}[t]
	\begin{center}
	\begin{subfigure}{0.32\columnwidth}
		\includegraphics[width=\linewidth, height=1.3in]{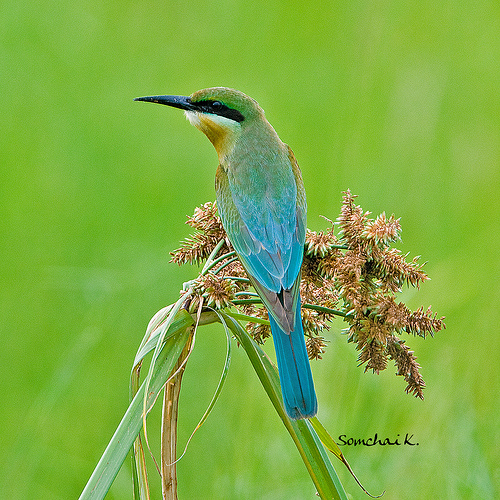}
		\caption{}
		\label{fig:1a}
	\end{subfigure}
	\begin{subfigure}{0.32\columnwidth}
		\includegraphics[width=\linewidth, height=1.3in]{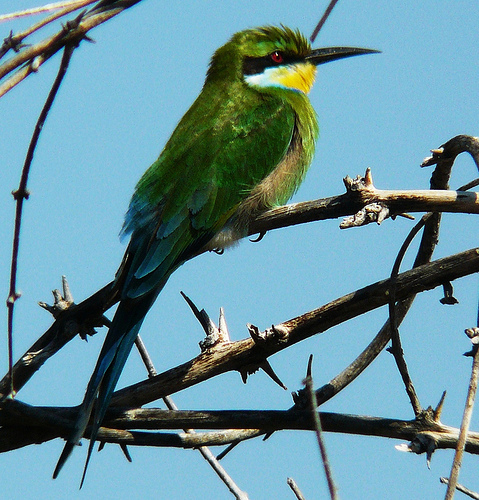}
		\caption{}
		\label{fig:1b}
	\end{subfigure}
	\begin{subfigure}{0.32\columnwidth}
	    \includegraphics[width=\linewidth, height=1.3in]{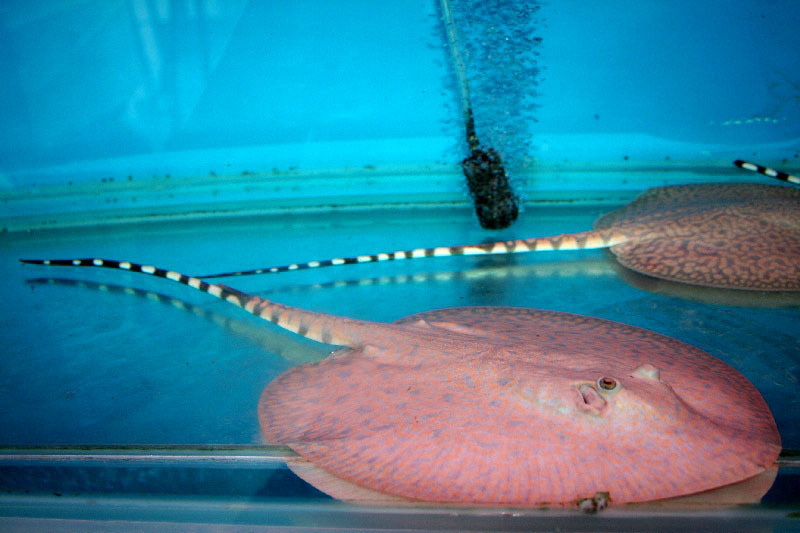}
	    \caption{}
	    \label{fig:1c}
	\end{subfigure}
	\end{center}
	\caption{Three samples from the ImageNet \cite{ILSVRC15} validation set. The First two images belong to the class hummingbird, whereas the third image is of an electric ray. In terms of the dominant color, the last two images are closer to each other. However, if we consider abstract features such as body patterns, pose or the type of animal, the first two images are closer to each other. Therefore, it is intuitive to learn different groups of features separately, in each layer of a network} 
	\label{fig:intro}
\end{figure}

In this view, processing an input according to its context is important because, in such a setting, models can extract different sets of features with varying priorities. To this end, networks that adaptively extract information have been introduced to converge to better optima in complex datasets, without merely increasing the number of parameters \cite{hu2017squeeze, srivastava2015highway, ha2016hypernetworks, convnet-aig, blockdrop}. These approaches utilize additional parametric functions that process the inputs of the model or each layer to accommodate the differences in data points while supporting the learning of the main task. However, they are still limited in the sense of utilizing a single column. 

The context of an input sample does not necessarily reflect the corresponding class but can provide more information, which may be present in multiple levels of abstractions. Figure \ref{fig:intro} shows three samples (Image \ref{fig:1a}, \ref{fig:1b} and \ref{fig:1c}) in the ILSVRC2012 \cite{ILSVRC15} validation set. Image \ref{fig:1a} and \ref{fig:1b} show hummingbirds sitting on a leafy or a thorny branch, with grass or sky as backgrounds, respectively. Image \ref{fig:1c} shows an electric ray immersed in water with a distinct body pattern and a pose that is different from the first two images. These contextual representations are distributed in multiple levels of a deep neural network. For instance, initial layers may capture the dominant color, the structure of edges and corners, whereas latter layers may capture more abstract information of the context such as the pose or even the class. Moreover, the image context which matters for the problem might be different from the visual contextual description provided by a human \cite{kahatapitiya2019context}. Simply put, the context of an input sample should be distributed throughout multiple layers of a network. 

In a parallel-path neural network, multiple groups of feature maps can be processed within a single layer. If each group contains a set of homogeneous feature maps that relates to a distinct context, the filters extracting features on the particular group can be learned to be specialized to the respective context. This enables the network to be more efficient, having a smaller set of dedicated filters per group. Such a smart usage of resources leads to improved performance at a given complexity in contrast to having a single large set of filters per layer. To do such a grouping of feature maps in each layer and to allocate inputs to the layer accordingly, we need an adaptive mechanism that routes between parallel sets of feature maps of two adjacent layers in a feature-dependent manner. 

Since the contextual representation of an input sample is distributed along with the depth of a network, it is important to have such mechanisms to allocate resources per each layer or layer segment according to the level of context represented by the incoming feature maps. Hence, two samples that are allocated to similar paths in the initial layers might get different path allocations in later layers. For example, Image \ref{fig:1b} and \ref{fig:1c} share similar background color domains which are different from Image \ref{fig:1a}. However, if body patterns and pose are considered, Image \ref{fig:1a} and \ref{fig:1b} share more similarities than Image \ref{fig:1c}. Therefore in such a multipath network, Image \ref{fig:1b} and \ref{fig:1c} might be processed along similar paths in initial layers even though they represent two distinct classes. In last layers, Image \ref{fig:1a} and \ref{fig:1b} may get allocated to similar paths. 

In this paper, we propose a multi-path architecture that consists of a novel adaptive cross-connecting mechanism between parallel sets of feature maps in successive layers of the network. Our scheme enables the parallel paths of the network to be specialized in distinct contexts while having a soft routing between them to process the inputs through context-dependent pathways. The weighting coefficients of the cross-connections are computed from the incoming sets of feature maps. We insert these cross-connections at selected locations along the depth of the network to perform a selective routing end-to-end. The outputs of such cross-connections act as inputs to the next layer or layer segment with dedicated families of filters per each parallel path. We summarize our main contributions to this paper as follows:
\begin{itemize}
    \item We introduce a feature-dependent cross-connecting mechanism between parallel paths of a multi-path network. The intuition here is to group homogeneous features of each layer into parallel paths and assign a soft routing between those paths, conditioned on the input. It allows the network to be more efficient, having paths specialized for different contexts and appropriate feature mixing between them. 
    \item Our models surpass equivalent baselines and state-of-the-art counterparts for network deepening, widening, and adaptive feature extraction with comparable complexity on image classification datasets. 
    \item We empirically show that the resource allocation between parallel paths of a layer is based on its level of context abstraction. In fact, the gates at initial stages route simple contexts such as colors, and gates at latter stages route more complex contexts such as types of objects.
\end{itemize}

\begin{figure*}[t]
\begin{center}
   \includegraphics[width=\linewidth]{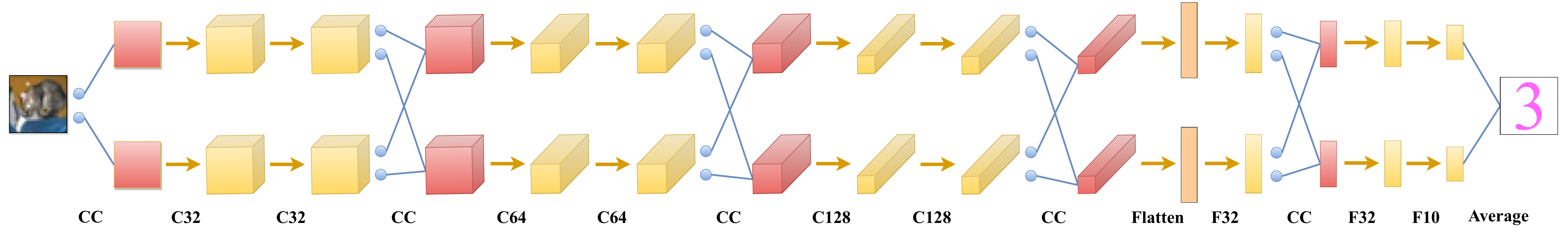}
\end{center}
\caption{Two-path CNN with adaptive cross-connections inserted at selected locations (referred to as BaseCNN-2 in the paper). C$n$ denotes a layer which carries parallel convolutional operations each with $n$ filters. F$n$ denotes a layer with parallel dense operations each with $n$ nodes. The yellow boxes represent the resulting tensors of such operations. The gated cross-connections (CC) are shown by blue lines and blue circles. The output tensors of such connections are shown in red. The input image is first expanded to two paths by gated connections, and the final layer responses are averaged to produce the end prediction}
\label{fig:basecnn2}
\end{figure*}

\begin{figure}[t]
\begin{center}
   \includegraphics[width=\linewidth]{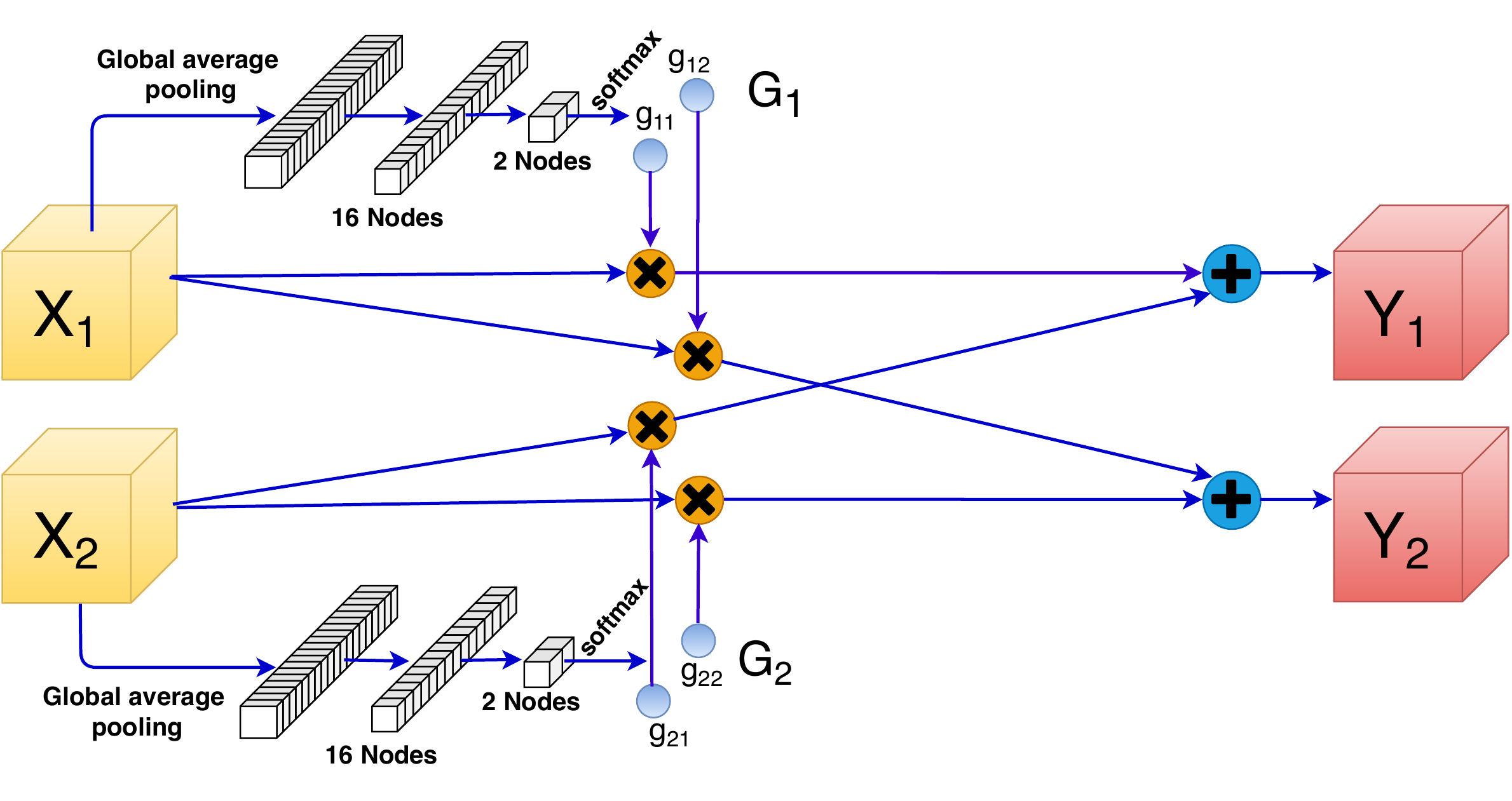}
\end{center}
\caption{Cross-Connections between two subsequent layers carrying 2 sets of feature maps each. The gating coefficients of the cross-connections are computed from the incoming sets of feature maps by a parametric computation.}
\label{fig:routing}
\end{figure}

\section{Background}
\label{se:related_work}
Neural network deepening \cite{alexnet,dsn, fitnet, vgg, resnet, preact-resnet} and widening \cite{wideresnet, inception} show promising improvements in complex datasets over conventional shallow neural networks \cite{rumelhart1986learning, lenet}. However, these methods do not effectively utilize the network capacity. Recent advances in adaptive learning, in which, the main task is supported by additional functions operating on input samples/feature maps of each layer, show a remarkable improvement over traditional deep or wide networks. These approaches either use additional learnable parametric functions to make the feature extraction of each layer sensitive to its input features \cite{hu2017squeeze, hu2018gather, wang2019eca, convnet-aig, srivastava2015highway}, or use latent sub-networks which predict the parameters of the main network based on a bulk of input samples \cite{ha2016hypernetworks}. Another set of approaches uses non-learnable computations between layers to regulate the flow of information dynamically \cite{sabour2017dynamic, emrouting}. Among these, our method is closely related to the use of feature-dependent layer-wise supportive functions.

SENet \cite{hu2017squeeze} introduced an adaptive re-calibration of feature maps by weighting them channel-wise, based on coefficients learned through a non-linear function corresponding to the particular set of feature maps. This is commonly referred to as channel-wise attention. Feature-dependent support has also been used for adjusting the depth of a network based on the input. In particular, both ConvNet-AIG \cite{convnet-aig} and BlockDrop \cite{blockdrop} introduced supportive algorithms to adaptively change the effective depth of Residual Networks \cite{resnet} per input sample. ConvNet-AIG uses a non-linear regularized parametric function in each residual block to produce hard gate values that decide whether to keep or drop the corresponding residual block, whereas BlockDrop takes this decision based on reinforcement learning. Highway networks \cite{srivastava2015highway, highway2} use additional gates computed on features maps to pass information across layers, without subjecting to attenuation. However, these approaches are limited to using a single column, whereas we use such feature-dependent support to allocate/route input features among parallel paths.

Approaches which use parallel convolution or residual operations in each layer \cite{inception, inception-v4, resnetxt}, or model ensembling \cite{alexnet,vgg} methods, do not allocate input samples or input feature maps to parallel operations in a data-dependent manner, and therefore, the parallel resource usage is not well optimized. The utility of parallel neural networks having each path conditioned on inputs pre-processed in different ways have proven to produce better results \cite{ciregan2012multi, wang2015multi}. However, these works do not introduce any connections between parallel neural networks, which is essential if we build a system to perform a soft assignment of paths in each layer, based on its level of context representation. To summarize, the utility of multi-path/column networks with feature-dependent resource allocation in each layer for learning a single task is not explored. 

Multi-path networks with cross-connections between hidden layers that carry parallel tensors are widely used in the common multitasking domain \cite{caruana1997multitask}, in which, multiple distinct tasks are performed on a given input sample. Cross-Stitch networks \cite{cross_stich} first proposed the use of cross-connections between parallel paths in selected layers of multi-path networks. However, the stitching coefficients which weigh the cross-connections are independently learned as conventional weights, and therefore, fixed for the whole dataset after training. Both Sluice Networks \cite{sluice} and NDDR-CNN \cite{nddr-cnn} build upon the intuition of Cross-Stitch Networks. 

Since the intention of Cross-Stitch Networks is to learn the fixed mix of task-specific and shared resources to handle multiple tasks per single input, the stitching coefficients being independently learned is sufficient. However, to build a multi-path network that performs adaptive path assignment based on the context of each hidden layer, so that the network can regulate the information flow end-to-end, these coefficients should be data-dependent. Therefore, we introduce a set of gating coefficients, generated as functions of the corresponding input tensors to the cross-connecting process. Our work can be interpreted as a fusion between the cross-stitching \cite{cross_stich} and channel-wise attention \cite{hu2017squeeze}, where the weighting coefficients of cross-connections are now learned by a set of non-linear, parametric, attention-like functions which operates conditioned on the input, in contrast to being learned individually.

\section{Adaptive Cross-Connections}
\label{se:systemarchitecture}
The adaptive cross-connecting algorithm which routes between two subsequent layers with parallel sets of feature maps says input and output layers of cross-connections, is a two-step process. First, each input tensor, i.e., the input set of feature maps, is fed into a non-linear parametric function to calculate gate probabilities/coupling coefficients which couple the corresponding tensor with each tensor in the output layer. Next, each tensor in the output layer is constructed based on a summation over all the input tensors weighted by the corresponding coupling coefficients. The coupling coefficients calculated as functions of the input tensors makes this a data-dependent process. 

Let's assume there are $m$ tensors [$\mathbf{X}_{i=1\dots m}$] in the input layer and $n$ tensors [$\mathbf{Y}_{j=1\dots n}$] in the output layer of a cross-connecting process. First, each tensor in the input layer predicts an $n$-dimensional vector $\mathbf{G}_i$, which contains $n$ probabilities which correspond to the coupling coefficients between tensor $i$ and each output tensor. $\mathbf{G}_i$ can be expressed as $[g_{i1}, \dots, g_{in}]$, where $g_{ij}$ corresponds to the scalar gate value between $\mathbf{X}_i$ and $\mathbf{Y}_j$.

The simplest way of calculating $\mathbf{G}_i$ is to directly perform a non-linear parametric computation on $\mathbf{X}_i$. However, this approach increases the number of parameters of this sub-network and the routing overhead, in turn. This increment is more evident if $\mathbf{X}_i$ is 3-dimensional. Therefore if $\mathbf{X}_i \in \mathbb{R}^{H\times W\times C}$, we feed $\mathbf{X}_i$ to a global average pooling layer to produce a $1\times 1\times C$ descriptor with channel-wise means. Since each channel in a convolutional feature map represents information extracted by a specific kernel, global average pooling compresses input while retaining important information related to the context. Each element in the output of pooling $\mathbf{Z}_i$ is given by, 
\begin{equation}
    \nonumber
	{(z_i)}_c = \frac{1}{H\times W}\sum_{a=1}^{H}\sum_{b=1}^{W}{(x_i)}_{a,b,c}.
	\label{eq:gap}
\end{equation}

$\mathbf{Z}_i$ is then fed into a non-linear computation to produce $n$ latent relevance scores $\mathbf{A}_i$ ($[a_{i1}, \dots, a_{in}]$). The non-linear function comprises of two fully-connected layers with 16 and $n$ number of nodes repectively, separated by $\mathrm{ReLU}$ activation. 
\begin{equation} \nonumber
	\mathbf{A}_i = \mathbf{W}_2(\mathrm{ReLU}(\mathbf{W}_1\mathbf{Z}_i))
	\label{eq:relevance}
\end{equation}
We impose $\mathrm{softmax}$ activation on top of the $n$ latent relevance scores $\mathbf{A}_i$ to calculate the gate probability vector $\mathbf{G}_i$. 
\begin{equation} \nonumber
    \mathbf{G}_i = \mathrm{softmax}(\mathbf{A}_i), \hspace{0.1in} i.e.,  \hspace{0.1in} g_{ij} = \frac{\mathrm{e}^{a_{ij}}}{\sum_{k=1}^{n} \mathrm{e}^{a_{ik}}} .
    \label{eq:G=softmaxA}
\end{equation}
This returns $n$ softmax scores which represent the probabilities of $\mathbf{X}_i$ being routed into each output ($\mathbf{Y}_{j=1,\dots,n}$). 

To construct $j^{\text{th}}$ output $\mathbf{Y}_j$, each input $\mathbf{X}_{i=1 \dots  m}$ is weighted by the corresponding scalar gate value ($g_{i=1\dots m,j}$), and summed over $i$. 
\begin{equation} \nonumber
	    \mathbf{Y}_j  =  \sum_{i=1}^{m} (g_{ij}\odot \mathbf{X}_i) 
\end{equation} 
Here $\odot$ stands for the element-wise multiplication between $g_{ij}$ and $\mathbf{X}_i$, which broadcasts the scalar $g_{ij}$ to match the dimensions of $\mathbf{X}_i$. 

We further show the cross-connecting process below in matrix form for simple illustration of pixel-wise operations. Let ${x_i}_{a,b,c}$ be the pixel value at location $(a,b,c)$ of $X_i$ and ${y_j}_{a,b,c}$ be the pixel value at location $(a,b,c)$ of $Y_j$. Let $\mathbf{G}$ be the $n\times m$ matrix with column $i$ denoting $n$ gate values computed from $\mathbf{X}_i$. The cross-connections between the two layers can be illustrated as,
\begin{equation}
    \begin{bmatrix}
    {y_1}_{a,b,c} \\
    \vdots \\
    {y_n}_{a,b,c}   \\
    \end{bmatrix}  = 
    \begin{bmatrix}
    g_{11} & \cdots &  g_{m1}   \\
    \vdots & \ddots & \vdots    \\
    g_{1n} & \cdots &  g_{mn} \\
    \end{bmatrix}
    \begin{bmatrix}
    {x_1}_{a,b,c} \\
    \vdots \\
    {x_m}_{a,b,c}   \\
    \end{bmatrix}.
    \label{eq:y_gx2}
\end{equation}

\begin{algorithm}[t]
	\caption{Routing between two adjacent layers with $m$ input and $n$ output sets of feature maps respectively.}
	\label{alg:routing}
	\begin{algorithmic}
		\STATE {\bfseries Input:} 
		\bindent
		\STATE $\mathbf{X}$: inputs \COMMENT{[$\mathbf{X}_i$ for $i=1,\dots,m$]}
		\eindent
		\STATE {\bfseries Calculating gate values:}
		\bindent
		\FOR{$i=1$ {\bfseries to} $m$}
		\STATE $\mathbf{Z}_i \leftarrow \mathrm{global\_average\_pooling}(\mathbf{X}_i)$
		\STATE $\mathbf{A}_i = [a_{i1},\dots ,a_{in}]  \leftarrow   
		\mathbf{W}^i_2(\mathrm{ReLU}(\mathbf{W}^i_1\boldsymbol{Z}_{i}))$
		\STATE $\mathbf{G}_i = [g_{i1},\dots ,g_{in}] \leftarrow \mathrm{softmax}(\mathbf{A}_i)$ 
		\ENDFOR	
		\eindent	
		\STATE {\bfseries Construction of outputs:}
		\bindent
		\FOR{$j=1$ {\bfseries to} $n$}
		\STATE $\mathbf{Y}_j \leftarrow  \sum_{i=1}^{m} (g_{ij}\odot \mathbf{X}_{i})$
		\ENDFOR
		\eindent	
		\STATE {\bfseries Return:} 
		\bindent
		\STATE $\mathbf{Y}$: outputs
		\COMMENT{[$\mathbf{Y}_j$ for $j=1,\dots,n$]}
		\eindent
	\end{algorithmic}	
\end{algorithm}

Cross-Stitch Networks \cite{cross_stich} use similar connections as in equation \ref{eq:y_gx2}. However, their coupling coefficients are learned independently, whereas we produce them based on input feature maps. Learning the coefficients independently is sufficient to produce a fixed mixture of shared and task-specific resources for a given dataset, in a scenario where multiple tasks are performed on the same input. In contrast, our intuition is to specifically make these coefficients feature-dependent and let the model decide the mixture of context-specific and shared resources corresponding to a given input sample.  

Figure \ref{fig:basecnn2} shows a two-path CNN with adaptive cross-connections inserted at selected positions. Figure \ref{fig:routing}, in detail, shows the cross-connecting process between two adjacent layers, each carrying two parallel sets of feature maps. Algorithm \ref{alg:routing} demonstrates the process of cross-connecting $m$ input tensors to $n$ output tensors. We insert these cross-connections at selected layers in multi-path networks to allow the parallel paths in the layers where there are no cross-connections to learn different context information independently. Since a cross-connecting layer only calculates a path assignment and does not directly contribute towards learning features for the main task, adding such layers does not increase the depth of the model.

\section{Experiments}
\label{se:experiments}
To validate the effectiveness of adaptive cross-connections, we conduct various experiments on both small and large scale image recognition datasets. In the ablation study conducted in CIFAR10 \cite{cifar100}, we compare our multi-path networks with baselines and evaluate the effect of the number of parallel paths. Next, we compare our multi-path architectures with existing state-of-the-art (SOTA) adaptive feature extractors in CIFAR10 and CIFA100 \cite{cifar100}. We further evaluate our models in ImageNet/ILSVRC2012 \cite{ILSVRC15} dataset to compare a two-path network with deeper and wider single path networks. We also use several visualization techniques on a multi-path CNN trained on a subset of ImageNet data to highlight the effect of adaptive path assignment.

\subsection{Ablation Study in CIFAR10}
\label{ss:ablation}

\begin{table}[t]
\caption{Compared Nets: $Cn$ - convolutional layer of $n$ filters. $Fn$ - fully connected layer of $n$ nodes.}
\label{tab:compared_networks}
\begin{center}
\renewcommand{\tabcolsep}{0.08mm}
\footnotesize
\begin{tabular}{@{}lr@{}}
\toprule
Network & Structure \\
\midrule
 BaseCNN &  $C32$ $C32$ $C64$ $C64$ $C128$ $C128$ $F32$ $F32$ $F10$ \\
 WideCNN &  $C64$ $C64$ $C128$ $C128$ $C256$ $C256$ $F32$ $F32$ $F10$ \\
 DeepCNN &  $C32$ $C32$ $C64$ $C64$ $C128$ $C128$ $C128$  \\& $C256$  $C256$ $C256$  $F32$ $F32$ $F10$ \\
 BaseCNN Ensemble &  Ensemble of 3 BaseCNNs \\
 All Ensemble &  Ensemble of BaseCNN, WideCNN and DeepCNN \\
 SENet &  SENet \cite{resnet} on BaseCNN and DeepCNN \\
 Cr-Stitch2 &  Cross-stitch network \cite{cross_stich} with 2 parallel BaseCNNs  \\
\midrule
 BaseCNN-X &  BaseCNN - X paths with adaptive cross-connections \\
 ResNet-X &  ResNet \cite{resnet} - X paths with adaptive cross-connections \\
\bottomrule
\end{tabular}
\end{center}
\end{table}

For the ablation study, the baselines are chosen as follows: BaseCNN is a standard convolutional neural network with 6 convolutional layers followed by 3 dense layers. DeepCNN is deeper with 10 convolutional layers. The total number of parameters in DeepCNN is still more than 3 times the parameters in BaseCNN. The WideCNN has a similar structure to BaseCNN, but in each convolutional layer, the number of filters is doubled. We also evaluate the performance of an ensemble of 3 BaseCNNs, and an ensemble of a BaseCNN, DeepCNN and a WideCNN.

We design our multi-path CNNs such that a single path is analogous to the BaseCNN. BaseCNN-X stands for a BaseCNN like a multi-path net with X number of parallel paths.  (Fig. \ref{fig:basecnn2} shows BaseCNN-2). The feature-dependent cross-connections are first used to  expand the input image to parallel RGB maps. More cross-connections are added after 2nd, 4th, 6th convolutional layers and after the 2nd dense layer. Finally, we average the final parallel dense layer predictions. In addition, we build SENets \cite{hu2017squeeze} on top of BaseCNN and DeepCNN with the SE operation added in each convolution. We also build a Cross-Stitch Network which comprises two BaseCNNs with non-adaptive stitching operations replacing the adaptive cross-connections. Table \ref{tab:compared_networks} briefly explains the compared structures.

\begin{table}[t]
	\caption{Ablation study in CIFAR10 - Classification errors (\%). BaseCNN-X surpasses single path deeper and wider networks and even ensembles of them. They also show superior performance to both SENets and Cross-stitch networks.}
	\label{tab:ablation} 
	\begin{center}
	\small
	\begin{tabular}[width=\columnwidth]{@{}lcr@{}}
		\toprule
 	    Network             & Params (M) & Error \%  \\
		\midrule 
		 BaseCNN/WideCNN/DeepCNN             &0.55/1.67/2.0 	&9.26/8.96/8.49  \\
	     BaseCNN Ensemble    &1.66 	&7.87  \\
		 All Ensemble        &4.27 	&6.9    \\
		 SEBaseCNN/SEDeepCNN           &0.58, 2.06 	&8.99,8.15    \\
		 Cr-Stitch2          &1.11 	&7.89    \\
		 VGG16 \cite{vgg}    &14.9 	&6.98  \\
		 \midrule 
		 BaseCNN-2          &1.11 	&7.03    \\
		 BaseCNN-3          &1.67 	&\textbf{6.51}    \\
		 BaseCNN-4          &2.22 	&\textbf{6.55}   \\
		\bottomrule
	\end{tabular}
	\end{center}
\end{table}

In the ablation study, all the models are trained for 200 epochs with a batch size of 128. We use an SGD optimizer with a momentum of 0.9 and an initial learning rate of 0.1, which is decayed by a factor of 10 after 80 and 150 epochs.  Table \ref{tab:ablation} illustrates the performance of baselines and our multi-path nets, where we report the least error obtained from 3 trials. All BaseCNN-X nets comfortably surpass the BaseCNN and even conventional deeper (DeepCNN) or wider (WideCNN) networks. BaseCNN-3 shows superior performance to the ensemble of 3 BaseCNNs and even the ensemble of all nets. This shows the promising nature of our adaptive cross-connections and further validates that the improvement is not merely due to the widened nature of our models. 

Our models also surpass SENets we built on top of BaseCNN and DeepCNN showing the superiority of feature dependent cross-connections of multi-path networks over deeper nets with the attentional recalibration of feature maps. In addition, BaseCNN-2 shows superior results to Cross-Stitch2 Net, confirming that adaptive cross-connections are more suitable for a parallel path network to learn from single-input-single-output distributions while adjusting the resource allocation per sample.

\subsection{Comparison with SOTA for Small-Scale Datasets}
\label{ss:sota_comparison}

\begin{table}[t]
	\caption{Classification error (\%) comparison with SOTA methods. ResNet20-3 outperforms ResNet110. ResNet20-3/4 and ResNet32-3/4 show on-par or superior performance to existing adaptive architectures which are mostly based on ResNet110. BaseCNN-X architectures surpass CNN based adaptive image classifiers which have similar or more number of parameters}
	\label{tab:sota}
	\begin{center}
	\renewcommand{\tabcolsep}{0.7 mm}
	\small
	\begin{tabular}[width=\columnwidth]{@{}lccr@{}}
		\toprule
 	    Network             & Params (M) & CIFAR10 &CIFAR100   \\
		\midrule 
		 ResNet20/110/164 \cite{resnet}                      &0.27/1.7/2.5 	&8.75/6.61/5.93    &-/26.88/25.16\\
		 WRN-40-2 \cite{wideresnet} &2.2   & 5.33    &26.04 \\
		 SEResNet110 \cite{hu2017squeeze}             &1.7   &5.21    &\textbf{23.85}\\
		 BlockDrop \cite{blockdrop}                   &1.7   &6.4  &26.3\\
	     ConvNet-AIG \cite{convnet-aig}               &1.78   &5.76   &-\\
	     ConvNet-AIG all\cite{convnet-aig}            &1.78   &5.14   &-\\
	     ResNet20-2 Ours       &0.55 	  &5.5    &27.36\\
		 ResNet20-3 Ours       &0.82     &5.18   &25.76\\
		 ResNet20-4 Ours       &1.1      &\textbf{4.96}   &\textbf{24.81}\\
		 ResNet32-2 Ours       &0.94 	 &5.14    &25.96\\
		 ResNet32-3 Ours       &1.41     &\textbf{4.96}   &\textbf{24.51}\\
		 ResNet32-4 Ours       &1.88     &\textbf{4.59}   &\textbf{23.52}\\
	     \midrule
		 Sabour \etal \cite{sabour2017dynamic}       &8.2    &10.6   &-\\
		 Highway \cite{srivastava2015highway, highway2} &2.3  &7.54   & -  \\
		 HyperNetworks \cite{ha2016hypernetworks}    &0.15 &7.23 &-  \\
		 BaseCNN-2/3/4  Ours     &1.11/1.67/2.22 &\textbf{6.53}/\textbf{6.09}/\textbf{6.26} &-\\
		\bottomrule
	\end{tabular}
	\end{center}
\end{table}

In this section, we compare our multi-path networks with existing deep, wide and adaptive image classifiers in the literature. We adopt our cross-connections to multi-path ResNets \cite{resnet} and modify the ResNet20 and ResNet32 variants with multiple paths. The initial convolution and pooling operation is shared which is followed by an adaptive one-to-many connector to expand to multiple parallel paths. We insert cross-connections after each set of residual blocks separated with strides and the final dense predictions are averaged. We report the performance of both ResNet and CNN variants of multi-path architectures in CIFAR10 and CIFAR100 datasets in Table \ref{tab:sota}. All the models are trained for 350 epochs where we used a batch size of 64 for ResNet variants and 128 for CNN variants. The initial learning rate of 0.1 is decayed by a factor of 10 after 150 and 250 epochs. We use standard data augmentation of pixel shift and random horizontal flip. We report the least error obtained from 3 trials.

Our multi-path ResNets which are based on ResNet20 and ResNet32 show superior performance to conventional ResNets. In particular ResNet20-3 (3 paths) is sufficient to surpass the deeper ResNet110's performance in both datasets. All our ResNet variants except ResNet20-2 surpass the Wide ResNet variant (WRN-40-2) \cite{wideresnet} which has 40 layers of depth, a width ratio of 2 and more complexity. Also, ResNet20-3/4 and ResNet32-3/4 architectures show on-par or superior performance to the adaptive networks such as SENet \cite{hu2017squeeze}, ConvNet-AIG \cite{convnet-aig} and BlockDrop \cite{blockdrop} which are based on ResNet110. ResNet20-3/4 and ResNet32-3 have significantly less number of parameters (0.82M, 1.1M, 1.41M) compared to ResNet110 and ResNet110 based adaptive networks (1.7M). The table also shows that the performances of CNN based adaptive image classifiers are surpassed by BaseCNN-X. 

It is also important to analyze the performance gain with the number of parallel paths. From Table \ref{tab:ablation} and Table \ref{tab:sota} we can observe that adding another parallel path to the BaseCNN gives a significant increment in the classification accuracy. However, the amount of accuracy increase is reduced with BaseCNN3, and BaseCNN4 shows no further improvement. Although the ResNet with 4 parallel paths gives the best performance, we can still observe the same phenomena as described above. In summary, given a dataset, the best accuracy gain is observed when extending a single path network to a dual-path network, and this gain is gradually reduced with the addition of more paths.

\subsection{ILSVRC2012 Dataset}
\label{ss:imagenet}

\begin{table}[t]
	\caption{Single-crop and 10-crop validation error (\%) in ILSVRC2012 dataset. ResNet18-2 with two paths comfortably outperforms ResNet18 and marginally outperforms ResNet34 and Wide ResNet18 with a width factor 1.5.}
	\label{tab:classi_img}
	\begin{center}
	\renewcommand{\tabcolsep}{1.2 mm}
	\small
	\begin{tabular}[width=\columnwidth]{@{}lccccr@{}}
		\toprule
 	    Network  & Params(M)  &\multicolumn{2}{c}{Single-Crop} & \multicolumn{2}{c}{10-Crop} \\
 	               & & Top-1 & Top-5 & Top-1 & Top-5  \\
		\midrule 
		 ResNet18 \cite{wideresnet} \cite{fbresnet}       & 11.7   &30.4 	&10.93    &28.22    &9.42 \\
		 ResNet34 \cite{wideresnet} \cite{resnet}   & 21.8  &26.77 	&8.77    &24.52    &7.46 \\
		 WRN-18-1.5 \cite{wideresnet}     & 25.9 &27.06 & 9.0 & &   \\
		 ResNet18-2 (Ours)                & 23.4 &\textbf{26.48} 	&\textbf{8.6}     &\textbf{24.5}    &\textbf{7.34}\\
		 \bottomrule
	\end{tabular}
	\end{center}
\end{table}

\begin{figure*}[t]
	\begin{center}
	\includegraphics[width=0.9\linewidth]{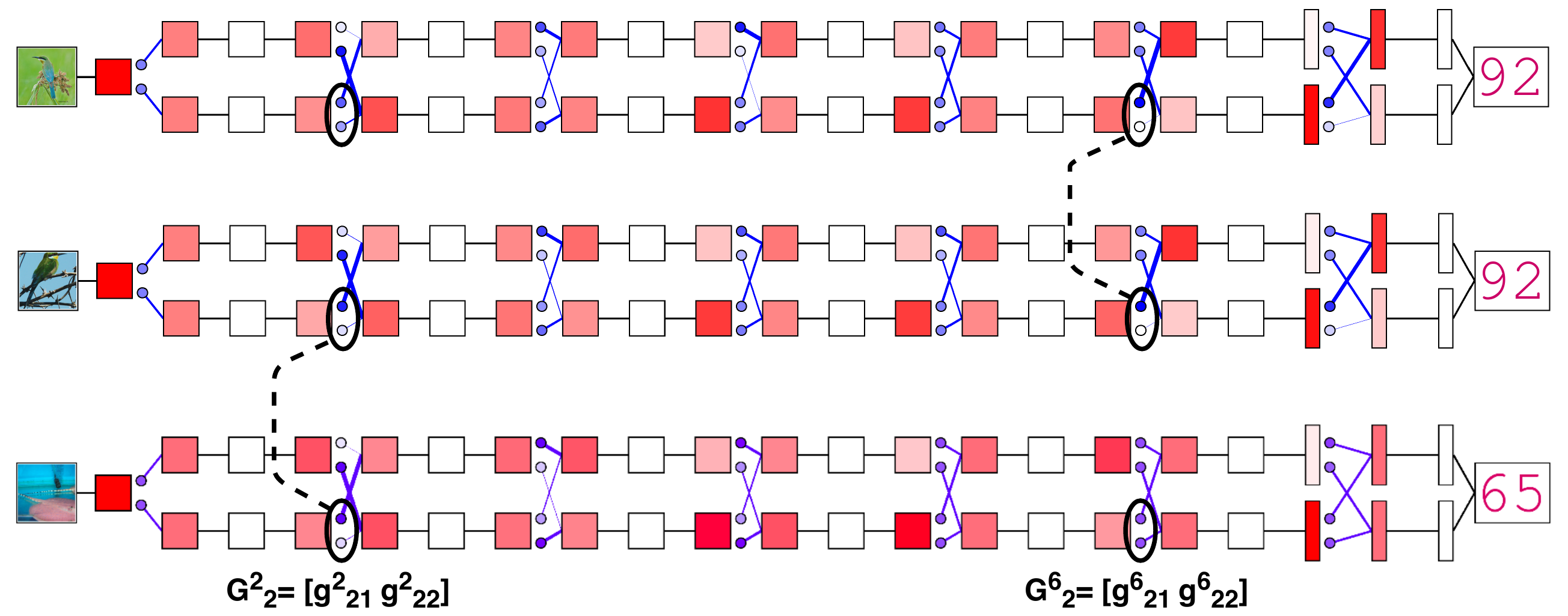}
	\end{center}
	\caption{Information flow of the input images in Figure \ref{fig:intro}. The relative activation strengths of the input and the output tensors of each cross-connecting layer are indicated by the saturation of red color. The contrast of blue circles and the thickness of the lines indicate gate strengths between paths. The gating vector $G^2_2$ at an initial stage of the network shows similar routing patterns for Image \ref{fig:1b} and \ref{fig:1c}, whereas the gating vector $G^6_2$ at a latter stage shows almost exactly the same routing for Image \ref{fig:1a} and \ref{fig:1b} (Best viewed on screen).}
	\label{fig:routes}
\end{figure*}

ILSVRC2012/ImageNet Dataset \cite{ILSVRC15} is a large-scale complex dataset with 1.3M training images and 50k validation images categorized into 1000 classes. To train on this dataset, we modify the ResNet18 variant \cite{resnet} with two paths (ResNet18-2). The insertion of cross-connections follows a similar setup as described in Sec. \ref{ss:sota_comparison} for ResNets. We train our model for 110 epochs with a batch size of 256 in two GPUs. We use an SGD optimizer with a momentum of 0.9 and an initial learning rate of 0.1, which is decayed by a factor of 10 once every 30 epochs. We augment images by re-scaling to 256$\times$ 256 and taking random crops of 224$\times$ 224, which are randomly flipped horizontally. 

Table \ref{tab:classi_img} shows the top-1 and top-5 errors evaluated for both single-crop and 10-crop testing in ILSVRC2012 dataset. ResNet18-2 comfortably surpasses ResNet18, and shows a slightly superior performance to ResNet34 which shares a similar amount of parameters to our design. Also, ResNet18-2 marginally surpasses the Wide ResNet18 \cite{wideresnet} with a width ratio of 1.5 (WRN-18-1.5). Due to the quadratic increment in the number of parameters with width enhancement, WRN-18-1.5 has more number of parameters than ResNet18-2.

\section{Visualizing and Understanding}
\label{se:visualization}
We use several visualization techniques to empirically validate the nature of the feature-dependency of our approach. Here, we train a two-path network based on VGG13 \cite{vgg}, namely VGG13-2, on a subset of ImageNet which consists of the first 100 classes. This subset contains 130k training images and 5000 validation images. Each path of the model is similar in structure to VGG13, but the number of convolutional filters in each layer is halved and the number of nodes in the dense layers is reduced to 256. The proposed cross-connections are added similarly to BaseCNN-2. 

\subsection{Information Flow through Cross-Connections}
\label{ss:routes}

We show the adaptiveness of information flow through our model by indicating the relative activation strengths of the parallel input and output tensors of each cross-connecting layer, and the corresponding gating strengths. To approximate the relative activation strength of each tensor, i.e., each path, we normalize its average activation w.r.t that of all tensors in a layer. The normalized activations are then mapped into red color intensities of the boxes representing the tensors in a cross-connecting layer. We represent the coupling coefficients, i.e., gate values, as blue circles connected with straight lines. The contrast of circles and the thickness of lines are proportional to the respective gate values. We only use this visualisation scheme on the layers with cross-connections. 

Figure \ref{fig:routes} shows the route visualizations of VGG13-2 for the three input images in Figure \ref{fig:intro}, which verifies the nature of the feature-dependency of our model. In particular, pay attention to the two gate vectors $G^2_2$ ([$g^2_{21}$ $g^2_{22}$]) and $G^6_2$ ([$g^6_{21}$ $g^6_{22}$]) which correspond to either initial or latter stages of the network respectively. Here, $G^l_i$ denotes the gate vector computed on input $i$ in cross-connecting layer $l$. Image \ref{fig:1a} and \ref{fig:1b} corresponding to the same class trigger different gating patterns in $G^2_2$, whereas Image \ref{fig:1b} and \ref{fig:1c} corresponding to different classes, show interestingly similar patterns ($g^2_{21}$ getting maximized). However, this behaviour is flipped in $G^6_2$, in which, Image \ref{fig:1a} and Image \ref{fig:1b} leads to equivalent gating patterns ($g^6_{21}$ getting maximized) which is distinguishably different from the Image \ref{fig:1c}.

\subsection{Maximizing Gate Activations}
\label{ss:act_max}

\begin{figure}[t]
	\begin{center}
	\begin{subfigure}{0.73\columnwidth}
		\includegraphics[width=\linewidth]{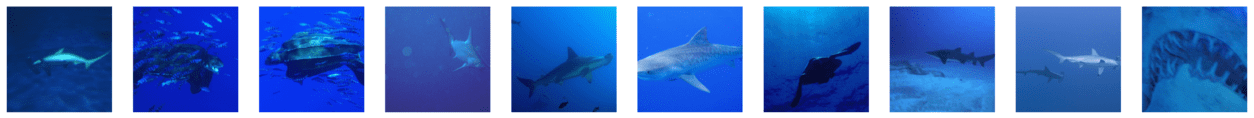}
		\includegraphics[width=\linewidth]{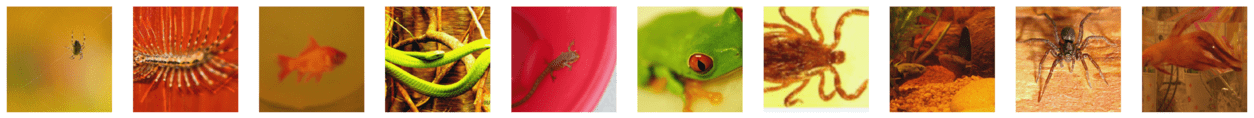}
		\caption{$g^2_{21}$}
	\end{subfigure}
	\begin{subfigure}{0.25\columnwidth}
			\includegraphics[width=\linewidth]{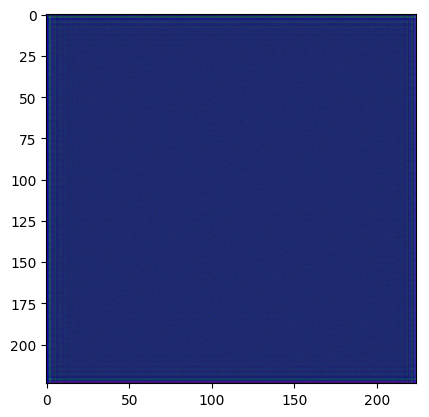}
	\end{subfigure}
	
	\begin{subfigure}{0.73\columnwidth}
		\includegraphics[width=\linewidth]{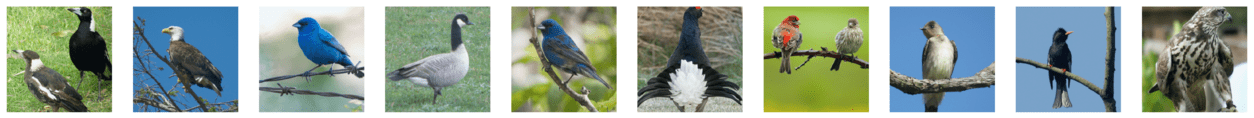}
		\includegraphics[width=\linewidth]{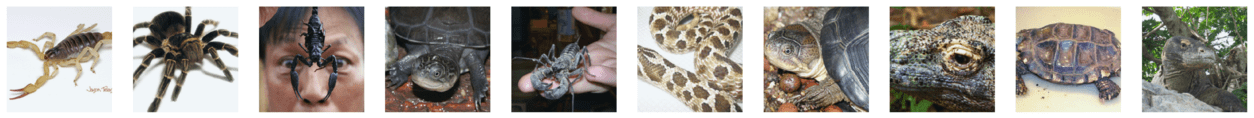}
		\caption{$g^6_{21}$}
	\end{subfigure}
	\begin{subfigure}{0.25\columnwidth}
			\includegraphics[width=\linewidth]{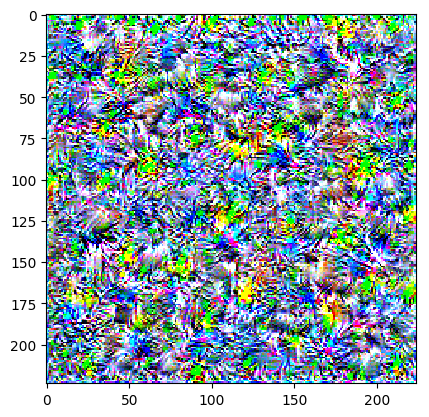}
	\end{subfigure}
	
	\end{center}
	\caption{In each subfigure, the top left row shows the 10 images with the highest activations and bottom left row shows the 10 images with the lowest activations for a particular gate, followed by a synthesized image that maximizes the gate neuron in the right. The gates in initial layers such as $g^2_{21}$ get triggered by low-level patterns (blue color), whereas the gates in deeper layers such as $g^6_{21}$ get triggered by more abstract patterns (bird poses/patterns).}
	\label{fig:gate_max}
\end{figure}

To further understand the described behaviour in Section \ref{ss:routes}, we plot the images from the validation set, which maximize and minimize the gates $g^2_{21}$ and $g^6_{21}$. If we consider each gate pair among [$g^2_{21}$, $g^2_{22}$] and [$g^6_{21}$, $g^6_{22}$], maximizing one gate value causes the other gate value to be minimized due to the softmax activation. Therefore, analysing one gate's two extremes lets us understand the contextual representation captured by the particular 2-dimensional gating vector. To further show the context which gives the maximum trigger at a particular gate, we freeze the trained network and synthesize an input which maximizes the corresponding gate neuron (before softmax, i.e., $a^2_{21}$ and $a^6_{21}$). We add L2 regularization to produce smooth images. This process is similar to class-specific image generation \cite{act_max}. 

Figure \ref{fig:gate_max} shows the results of this study. In each  sub-figure, the top left row shows the 10 images with highest activation corresponding to each gate. The bottom left row shows the 10 images with the lowest gate activation. In the right, the synthesized input which maximized the gate neuron is shown. Since $G^2_2$ corresponds to initial stages of the network, its component $g^2_{21}$ gets triggered by low-level, i.e., less-abstract features. The images which maximally activate this gate frequently contain an overall blue color. Correspondingly, the synthesized image that gives the maximum trigger at $g^2_{21}$ can be identified as a uniformly distributed blue image with minimal pattern complexity. In contrast, the gating vector $G^6_2$ corresponds to the latter stages of the network, and therefore, the corresponding gate $g^6_{21}$ gets triggered for more abstract features. Here, all 10 images with the highest activations are associated with bird poses. The synthesized image further shows complex patterns possibly related to birds. In both cases, the images which activate the particular gate to the lowest extent shows a minimal match with these patterns.

Based on these visualizations, we can describe the behaviour of gating patterns presented in Section \ref{ss:routes}. The blue backgrounds of Image \ref{fig:1b} and \ref{fig:1c} have been prominent causing them to maximally trigger $g^2_{21}$. However in latter stages of the network, $g^6_{21}$ is triggered by Image \ref{fig:1a} and \ref{fig:1b} as they both correspond to birds with similar poses. This verifies that the context information which affects the gating mechanism has changed to more abstract features similar to the concept of a class. This further emphasizes that the image context which matters in the problem domain is distributed among multiple layers of the network, scaling from low-level to high-level depending on the depth. Therefore, having cross-connections at multiple depths enables gating, i.e., path allocation, based on the level of context abstraction represented at the corresponding depth.

\subsection{Weight Histograms of Parallel Paths}
\label{ss:weight_hist}

\begin{figure}[t]
	\begin{center}
	\begin{subfigure}[b]{0.32\columnwidth}
		\includegraphics[width=\linewidth]{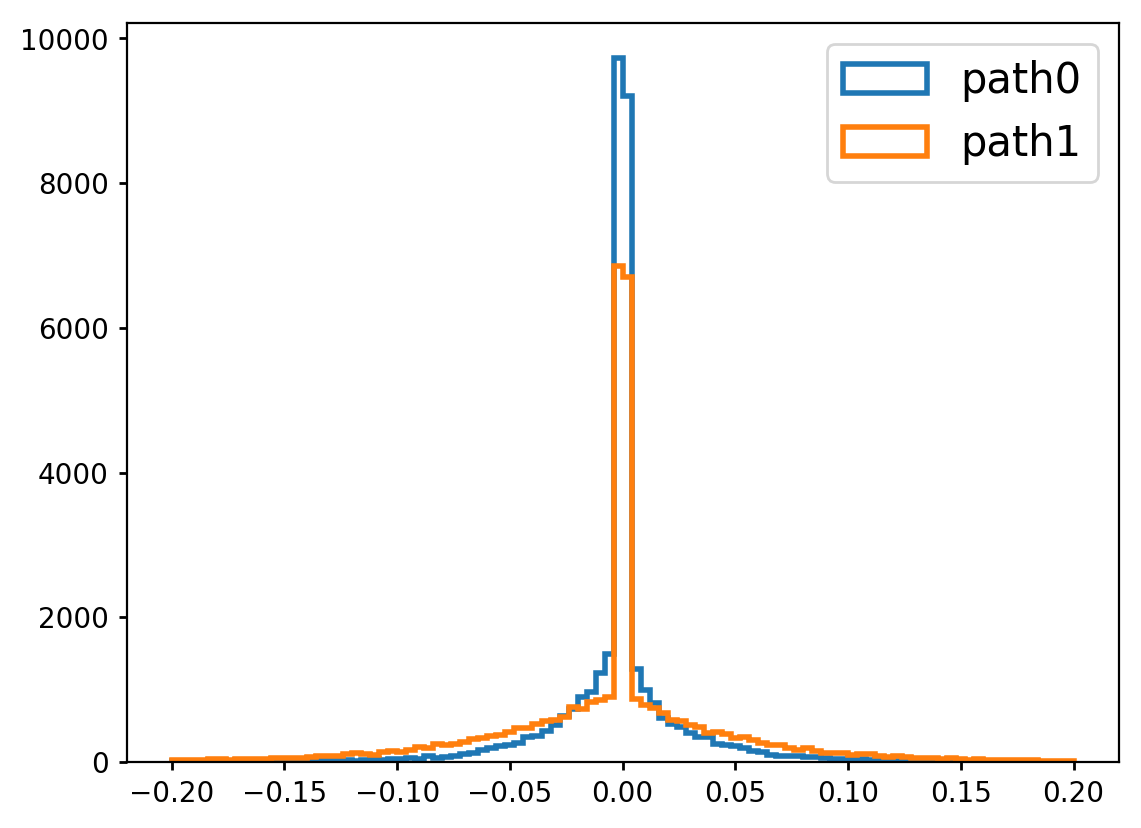}
		\caption{layer4}
	\end{subfigure}
	\begin{subfigure}[b]{0.32\columnwidth}
		\includegraphics[width=\linewidth]{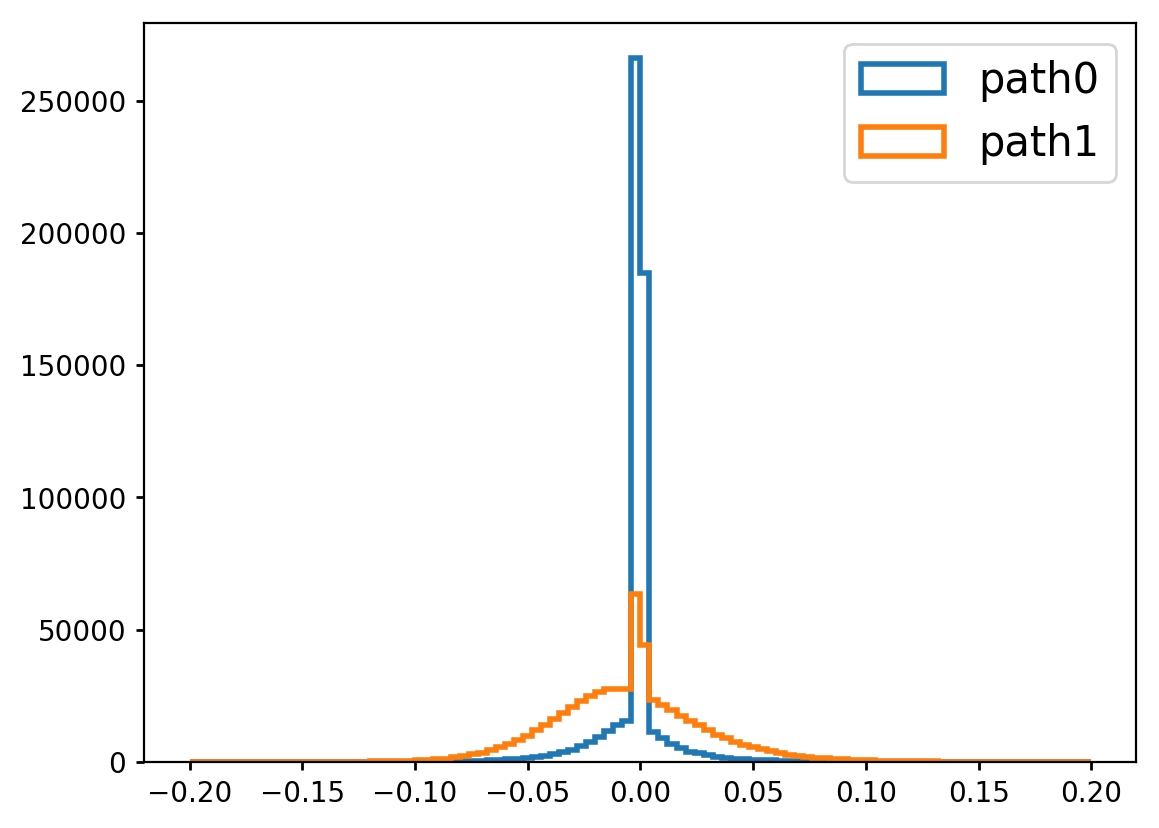}
		\caption{layer8}
	\end{subfigure}
	\begin{subfigure}[b]{0.32\columnwidth}
		\includegraphics[width=\linewidth]{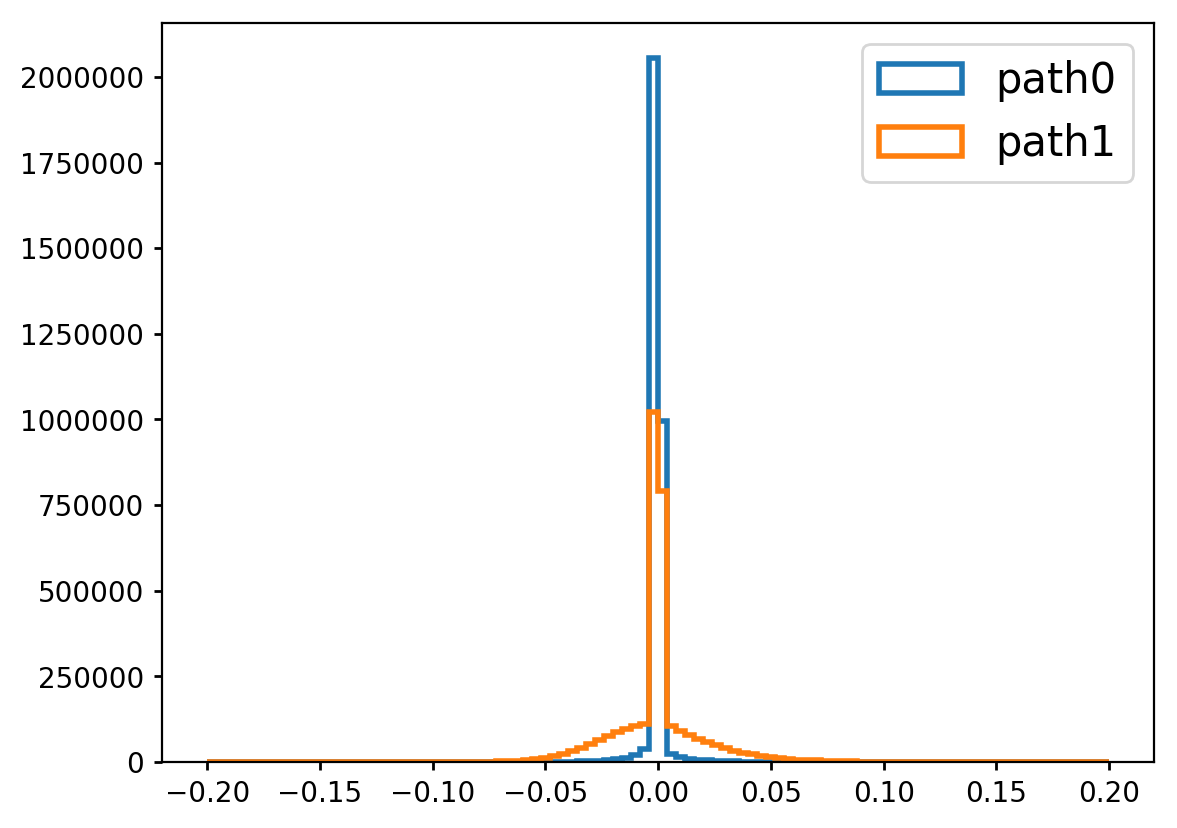}
		\caption{layer11}
	\end{subfigure}
	\end{center}
	\caption{Weight histograms of parallel paths at selected layers. By grouping homogeneous feature maps to parallel paths, the set of filters in each path has become distinct from each other.}
	\label{fig:weight_hist}
\end{figure}

One of our motivations for deploying a multi-path architecture is to group homogeneous feature maps into parallel paths, so that, each set of filters corresponding to a path can be dedicated to its context. This implies that the set of filters in each parallel path should learn distinct features, reducing the redundancy. To validate this claim, we plot weight histograms of parallel paths at selected layers. Figure \ref{fig:weight_hist} shows these weight histograms of parallel paths at feed-forward layers $4$, $8$ and $11$. In these cases, we can see that the weight histograms of parallel paths have become distributed.

\section{Conclusion}
\label{se:conclusion}
In this paper, we explored the utility of multi-path neural networks as dynamically adaptive models to learn from complex datasets. In particular, we selectively introduced adaptive cross-connections between successive pairs of layers in a multi-path network, to cluster similar feature maps into parallel paths and learn a soft routing between them. The proposed cross-connections are weighted by a set of non-linear parametric coefficients produced based on incoming sets of feature maps to make the path selection process feature-dependent. Simply put, in the forward pass, an input image to this multipath network gets adaptively allocated among parallel paths in each layer, based on the nature of the feature maps in the corresponding layer. The experiments conducted on both small and large-scale image classification datasets show that such multi-path networks are capable of surpassing state-of-the-art adaptive image classifiers and conventional single path networks of increased width or depth which are of similar or even higher complexity. We further validate the nature of the feature dependency of our model and its ability to capture and adapt to the context of an input which is distributed among multiple layers along with the depth. 

\section{Acknowledgement}
\label{se:ack}
We thank Sanath Jayasena for arranging insight discussions which supported this work.


\balance
\bibliographystyle{IEEEtran}
\bibliography{egbib}

\end{document}